\documentclass[runningheads]{llncs}
\usepackage{graphicx}
\usepackage{amsmath,amssymb}
\usepackage[utf8]{inputenc}
\usepackage{enumitem}
\usepackage{cite}
\usepackage{algorithm}
\usepackage[noend]{algpseudocode}
\usepackage{hyperref}
\usepackage{amsfonts}
\usepackage{amssymb,amsmath,bm,bbm}
\usepackage{multirow}
\usepackage[font=small,labelfont=bf,tableposition=top]{caption}
\usepackage[table]{xcolor}
\DeclareCaptionLabelFormat{andtable}{#1~#2  \&  \tablename~\thetable}
\usepackage{graphicx}
\usepackage{subfig}

\DeclareMathOperator*{\argmin}{argmin}

\def\vw{{\bm{w}}}
\def\vx{{\bm{x}}}
\begin{document}
\title{An Integrated Approach to Produce Robust Deep Neural Network Models with High Efficiency \thanks{This work was partly supported by NSF Grants DMS-1854434, DMS-1924548, DMS-1952644, DMS-1924935, and DMS-1952339.}}
\author{Zhijian Li\inst{1} \and
Bao Wang\inst{2} \and
Jack Xin\inst{1}}
\authorrunning{Z. Li et al.}
\institute{University of California, Irvine, CA 92697, USA \and
The University of Utah , Salt Lake City, UT 84112, USA\\
\email{\{zhijil2,jack.xin\}@uci.edu, wangbaonj@gmail.com}}
\maketitle              % typeset the header of the contribution
\begin{abstract}
Deep Neural Networks (DNNs) need to be both efficient and robust for practical uses. Quantization and structure simplification are promising ways to adapt DNNs to mobile devices, and adversarial training is one of the most successful methods to train robust DNNs. In this work, we aim to realize  both advantages
by applying a convergent relaxation quantization algorithm, i.e., Binary-Relax (BR), to an adversarially trained robust model, 
i.e. the ResNets Ensemble via Feynman-Kac Formalism (EnResNet). We discover that high-precision quantization, such as ternary (tnn) or 4-bit, produces sparse DNNs. However, this sparsity is unstructured under adversarial training. To solve the problems that adversarial training jeopardizes DNNs' accuracy on clean images and break the structure of sparsity, we design a trade-off loss function that helps DNNs preserve natural accuracy and improve channel sparsity. With our newly designed trade-off loss function, we achieve both goals with no reduction of resistance under weak attacks and very minor reduction of resistance under strong adversarial attacks. Together with our model and algorithm selections and loss function design, we provide an integrated approach to produce robust DNNs with high efficiency and accuracy.  Furthermore, we
provide a missing benchmark on robustness of quantized models.

\end{abstract}
\section{Introduction}

\subsection{Background}
Deep Neural Networks (DNNs) have achieved significant success in computer vision and natural language processing. Especially, the residual network (ResNet)\cite{resnet} has
achieved remarkable
%state-of-the-art 
performance on image classification and has become one of the most important neural network architectures in the current literature. 

Despite the tremendous 
%large 
success of DNNs, researchers still try to strengthen two properties of DNNs, robustness and efficiency. In particular, for security-critic and on the edge applications. Robustness keeps the model accurate under small adversarial perturbation of input images, and efficiency enables us to
fit DNNs into embedded system, such as smartphone. Many adversarial defense algorithms
%defensive 
%methods 
\cite{PGD, Guo, Alexey,Zhang} have been proposed
%invented 
to improve
%increase 
the robustness of DNNs.
%Neural Network. 
Among them, adversarial training is one of the most effective and powerful methods. On the other hand, quantization \cite{bc, xnor}, producing models with low-precision weights, and structured simplification, such as channel pruning \cite{he2017channel, zhuang2018discrimination}, are promising ways to make models computationally efficient. 
\section{Related Work}
\subsection{Binary Quantization}
Based on the binary-connect (BC)
%BC, 
\cite{Yin} proposed an 
%a 
improvement of BC called Binary-Relax, which makes the weights converge to the binary weights
%binary 
from the floating point weights
%float 
gradually. Theoretically, \cite{Yin} provided the convergence analysis of BC, and \cite{Hao} presented an ergodic error bound of BC.
The space of m-bit quantized weights $\mathcal{Q}\subset \mathbb{R}^n$ is a union of disjoint one-dimensional subspaces of $R^n$. When $m\geq 2$, it is formulated as:

$$\mathcal{Q}=\mathbb{R}_{+}\times\{0, \pm 1, \pm 2,\cdots, \pm 2^{m-1}\}^n=\bigsqcup_{l=1}^p\mathcal{A}_l$$
where $\mathbb{R}_{+}$ is a float scalar. For the special binary case of $m=1$, we have $\mathcal{Q}=\mathbb{R}_{+}\times\{\pm 1\}^n$. \\
We minimize our objective function in the subspace $\mathcal{Q}$. Hence, the problem of binarizing weights can be formulated in the following two equivalent forms:
\begin{itemize}
\item I.
$$\argmin_{u\in \mathcal{Q}}\mathcal{L}(u)$$
\item II.
\begin{equation}
\argmin_{u\in \mathbb{R}^n}\mathcal{L}(u)+\chi_{\mathcal{Q}}(u)\label{eq1}\text{\ \  where \ \ }
 \chi_{\mathcal{Q}}(u) = \begin{cases} 
          0 & u\in \mathcal{Q} \\
         \infty & else
       \end{cases}
    \end{equation}    
\end{itemize}
Based on the alternative form II, \cite{Yin} relaxed the optimization problem to:
\begin{equation}
    \argmin_{u\in \mathbb{R}^n}\mathcal{L}(u)+\frac{\lambda}{2}dist(u,\mathcal{Q})^2\label{eq2}
\end{equation}
Observing \eqref{eq2} converges to \eqref{eq1} as $\lambda \to \infty$, \cite{Yin} proposed a relaxation of BC:
\[\begin{cases} 
          w_{t+1}=w_{t}-\gamma \nabla \mathcal{L}_t(u_t) \\
         u_{k+1}=\argmin_{u\in \mathbb{R}^n}\frac{1}{2}\|w_{t+1}-u\|^2+\frac{\lambda}{2}dist(u,\mathcal{Q})^2
       \end{cases}
    \]
It can be shown that $u_{k+1}$ above has a closed-form solution
$$u_{t+1}=\frac{\lambda\text{proj}_{\mathcal{Q}}(w_{t+1})+w_{t+1}}{\lambda+1}$$
\iffalse
\begin{algorithm*}
\caption{Binary-Relax quantization algorithm}\label{euclid}
\begin{algorithmic}[1]
\State\textbf{Input:} mini-batches \{$(\mathbf{x}_1,\mathbf{y}_1),\cdots (\mathbf{x}_m,\mathbf{y}_m)$\}, $\lambda_0=1$, growth rate $\rho>1$, learning rate $\gamma$, initial float weight $w_0$, initial binary weight $u_0=w_0$, cut-off epoch $M$
\State\textbf{Output:} a sequence of binary weights $\{u_t\}$
\For{$t=1,\cdots, N$}\Comment{N is the number of epochs}
\If{$t<M$}
\For{$k=1,\cdots, m$}
\State $w_{t}=w_{t-1}-\gamma_t \nabla_k \mathcal{L}(u_{t-1})$
\State $u_t=\frac{\lambda_t\cdot \text{Proj}_{\mathcal{Q}}(w_t)+w_t}{\lambda_t+1}$
\State $\lambda_{t+1}=\rho\cdot \lambda_t$
\EndFor
\Else
\For{$k=1,\cdots, m$}
\State $w_{t}=w_{t-1}-\gamma_t \nabla \mathcal{L}(u_{t-1})$
\State $u_t=\text{Proj}_{\mathcal{Q}}(w_t)$ \Comment{This is precisely Binary-Connect}
\EndFor
\EndIf
\EndFor
\State \textbf{return} quantized weights $u_N$
%\label{alg:br}
%\EndProcedure
\label{alg:br}
\end{algorithmic}
\end{algorithm*}
\fi
\subsection{Adversarial Attacks}
As \cite{szegedy2013intriguing} discovered the limited continuity of DNNs' input-output mapping, the outputs of DNNs can be changed by adding 
%fairly small 
imperceptible adversarial
perturbations to input data. The methods that generate perturbed data can be
%are 
adversarial attacks, and the generated perturbed data are called adversarial examples. In this work, we focus on three benchmark adversarial attacks Fast gradient sign method (FGSM) \cite{fgsm}, iterative FGSM (IFGSM), and Carlini and Wagner method (C\&W) \cite{cw}. In this study, we denote attacks FGSM, IFGSM, and CW to be $A_1$, $A_2$, and $A_3$ respectively.

\subsection{Adversarial Training}
\cite{benchmark} rigorously established a rigorous benchmark to evaluate the robustness of machine learning models and investigated almost all current popular adversarial defense algorithms.
They conclude that adversarially trained models are more robust models with other types of defense methods. \cite{Zhang} also shows that adversarial training is more powerful than other methods such as gradient mask and gradient regularization \cite{Alexey}. While a natural training has objective function:
\begin{equation}
    \mathcal{L}(\omega)=\frac{1}{N}\sum_{i=1}^{N}l\big(f(w,\bold{x}_i),y_i\big)
    \label{eq:nat}
\end{equation}
Adversarial training \cite{Anish, PGD} generates perturbed input data and train the model to stay stable under adversarial examples. It has the following objective function:
\begin{equation}
    \mathcal{L}(w)=\frac{1}{N}\sum_{n=1}^{N}\max_{\tilde{x}_n\in D_n}l(f(w,\tilde{x}_n),y_n)\label{eq:ad}
\end{equation}
where $D_n=\{x|\|x-x_n\|_{\infty}<\delta \}$. $l$ and $f$ are the loss function and the DNN respectively. In this work, we denote (\ref{eq:nat}) to be $\mathcal{R}_{nat}$ and (\ref{eq:ad}) to be $\mathcal{R}_{rob}$.
A widely used method to pratically find $\tilde{x}_n$ is the projected gradient descent (PGD) \cite{PGD}.  \cite{Sinha} investigated the properties of the objective function of the adversarial training, and \cite{Yisen} provided convergence analysis of adversarial training based on the previous results.

%\subsection{Ensemble of ResNets via the Feynman-Kac Formula}
\textbf{Feynman-Kac formalism principled Robust DNNs:} 
Neural ordinary differential equations (ODEs) \cite{chen2018neural} are a class of DNNs that use an ODE to describe the data flow of each input data. Instead of focusing on modeling the data flow of each individual input data, \cite{enresnet,wang2018deep,li2017deep} use a transport equation (TE) to model the flow for the whole input distribution. In particular, from the TE viewpoint, \cite{enresnet} modeled training ResNet as finding the optimal control of the following TE
\begin{eqnarray}
\label{PDE-Eq7}
\begin{cases}
\frac{\partial u}{\partial t} (\vx, t)+ G(\vx, \vw(t))\cdot \nabla u (\vx, t)= 0,\ \ \vx\in \mathbb{R}^d,\\
 u(\vx, 1) = g(\vx),\ \ \vx \in \mathbb{R}^d,\\
 u(\vx_i, 0) = y_i, \ \ \vx_i\in T,\ \ \mbox{with}\ T\ \mbox{being the training set.}
\end{cases}
\end{eqnarray}
where $G(\vx, \vw(t))$ encodes the architecture and weights of the underlying ResNet, $u(\vx, 0)$ serves as the classifier, $g(\vx)$ is the output activation of ResNet, and $y_i$ is the label of $\vx_i$.

Based on equation (\ref{PDE-Eq7}), \cite{enresnet} interpreted adversarial vulnerability of ResNet as arising from the irregularity of $u(\vx, 0)$ of the above TE. %model. 
To enhance $u(\vx, 0)$'s regularity, they added a diffusion term, $\frac{1}{2}\sigma^2\Delta u(\vx, t)$, to the governing equation of \eqref{PDE-Eq7} which results to the convection-diffusion equation (CDE). By the Feynman-Kac formula, $u(\vx, 0)$ of the CDE can be approximated by the following two steps:
\begin{itemize}[leftmargin=*]
\item Modify ResNet by injecting Gaussian noise to each residual mapping.
\item Average the output of $n$ jointly trained modified ResNets, and denote it as En$_n$ResNet.
\end{itemize}
\cite{enresnet} have noticed that EnResNet, comparing to ResNet with the same size, can significantly improve both natural and robust accuracies of the adversarially trained DNNs. In this work, we leverage the robust advantage of EnResNet to push the robustness limit of the quantized adversarially trained DNNs. 

\textbf{TRADES:} It is well-known that adversarial training will significantly reduce
%jeopardize 
the accuracy of models on clean images. For example, ResNet20 can achieve about 92\% accuracy on CIFAR10 dataset. However, under the PGD training using $\epsilon=0.031$ with step-size $0.007$
%0.031 
and the number of iterations 10, it only has 76\% accuracy on clean images of CIFAR10. \cite{Zhang} designed a trade-off loss function, TRADES, that balances the natural accuracy and adversarial accuracy. The main idea of TRADES is to minimize the difference of adversarial error and the natural error:
 %TRADES:
\begin{equation}
\label{eq:trades}
    \text{min}_f\mathbb{E}\Big(l(f(X), Y)+\text{max}_{X'\in \mathbb{B}(X,\epsilon)}l\big(f(X), f(X')\lambda \big)\Big)
\end{equation}
TRADES is considered to be the state-of-the-art adversarial defense method, it outperforms most defense methods in both 
robust accuracy and natural accuracy. \cite{benchmark} investigated various defense methods, and TRADES together with PGD training outperforms all other methods they tested. In this work, we will design our own trade-off function that works with PGD training. TRADES will provide important baselines for us.  

\section{Quantization of EnResNet}
We propose a framework that integrates quantization and adversarial training, which is shown in Algorithm \ref{alg:br}. We know that the accuracy of a quantized model will be lower than its counterpart with floating point weights
%float equivalent 
because of loss of precision. However, we want to know that whether a quantized model is more vulnerable than its float equivalent under adversarial attacks? In this section, we study this question by comparing the accuracy drops of the natural accuracy and robust accuracy from float weights to binary weights. Meanwhile, we also investigate the performances between two quantization methods BC and BR.  
\begin{algorithm}
\caption{Binary-Relax quantization algorithm for adversarial training}\label{euclid}
\begin{algorithmic}[1]
\State\textbf{Input:} mini-batches \{$(\mathbf{x}_1,\mathbf{y}_1),\cdots (\mathbf{x}_m,\mathbf{y}_m)$\}, $\lambda_0=1$, growth rate $\rho>1$, learning rate $\gamma$, initial float weight $w_0$, initial binary weight $u_0=w_0$, cut-off epoch $M$
\State\textbf{Output:} a sequence of binary weights $\{u_t\}$
\For{$t=1,\cdots, N$}\Comment{N: the number of epochs}
\For{k = 1,$\cdots$, m}
\For{n = 1,$\cdots$, 10}
\State $\mathbf{x}_k^{(n)}=\mathbf{x}_k^{n-1}+\alpha\cdot \text{sign} \big( \nabla_u\mathcal{L}(f(u_{t-1},\mathbf{x}_k^{(n-1)}),\mathbf{y}_k)\big)$ 
\EndFor
\State $\mathbf{x}_k'=\mathbf{x}_k+\text{Clip}_{\epsilon}(\mathbf{x}_k^{(n)})$ \Comment{Perturb training data}
\EndFor
\If{$t<M$}
\For{$k=1,\cdots, m$} \Comment{m: number of mini-batch}
\State $w_{t}=w_{t-1}-\gamma_t \nabla_u \mathcal{L}\big(f(u_{t-1}, \mathbf{x}_k'), \mathbf{y}_k\big)$
\State $u_t=\frac{\lambda_t\cdot \text{Proj}_{\mathcal{Q}}(w_t)+w_t}{\lambda_t+1}$
\State $\lambda_{t+1}=\rho\cdot \lambda_t$
\EndFor
\Else
\For{$k=1,\cdots, m$}
\State $w_{t}=w_{t-1}-\gamma_t \nabla \mathcal{L}(u_{t-1})$
\State $u_t=\text{Proj}_{\mathcal{Q}}(w_t)$ \Comment{This is precisely Binary-Connect}
\EndFor
\EndIf
\EndFor
\State \textbf{return} quantized weights $u_N$
%\label{alg:br}
%\EndProcedure
\end{algorithmic}
\label{alg:br}
\end{algorithm}

\subsection{Experimental Setup}
\textbf{Dataset.} We use one of the most popular datasets CIFAR-10 \cite{cifar} to evaluate the quantized models, as it would be convenient to compare it with the float models used in \cite{enresnet,Zhang}.

\textbf{Baseline.} Our baseline model is the regular ResNet. To our best knowledge, there has been work done before that investigated the robustness of models with quantized weights, so we do not have an expected accuracy to beat. Hence, our goals are to compare the robustness of binarized ResNet and binarized EnResNet and to see how close the accuracy of quantized models can be to the float models in \cite{enresnet, Zhang}.

\textbf{Evaluation.} We evaluate both natural accuracy and robust accuracy for quantized advarsarial trained models. We examine the robustness of models FGSM ($A_1$),  IFGSM ($A_2$), and C\&W ($A_3$). In our recording, $N$ denotes the natural accuracy (accuracy on clean images) of models. For FGSM, we use $\epsilon=0.031$ as almost all works share this value for FGSM. For  IFGSM, we use $\alpha=1/255$, $\epsilon=0.031$, and number of iterations 20. For C\&W, we have learning rate 0.0006 and the number of iterations 50.

\textbf{Algorithm and Projection} We set BC as our baseline algorithm, and we want to examine that whether the advantage of the relaxed algorithm \ref{alg:br} in \cite{Yin} is preserved under adversarial training. In both BC and BR, we use the wildly used binarizing projection proposed by \cite{xnor}, namely:
$$\text{Proj}_\mathcal{Q}(w)=\mathbb{E}[|w|]\cdot sign(w)=\frac{||w||_1}{n}\cdot sign(w)$$
where $sign(\cdot)$ is the component-wise sign function and $n$ is the dimension of weights. 
\subsection{Result}
First, we verify that the Ensemble ResNet consistently outperforms ResNet when binarized. We adversarially train two sets of EnResNet and ResNet with the similar number of parameters. As shown in table \ref{table 2}, EnResNet has much higher robust accuracy for both float and quantized models.

Second, we investigate the performances of Binary-Connect method (BC) and Binary-Relax method (BR). We verify that BR outperforms BC (Table \ref{tab:binarize}). A quantized model trained via BR provides higher natural accuracy and robust accuracy. As a consequence, we use this relaxed method to quantize DNNs in all subsequent experiments in this paper.\\
\begin{table}[ht!]
    \centering
    \begin{tabular}{c c@{\hspace{1.5\tabcolsep}}
    c@{\hspace{1.5\tabcolsep}} 
    c@{\hspace{1.5\tabcolsep}}
    c @{\hspace{1.5\tabcolsep}}
    c}
         \hline\hline
    net(\#params) & model &$N$ & $A_1$ & $A_2$ & $A_3$\\
    \hline
   En$_1$ResNet20 (0.27M)
    &BR &\bf{69.60}\% & \bf{47.17}\% & \bf{43.89}\% & \bf{58.79}\%\\
   ResNet20 (0.27M)
    &  BR& 66.81\% & 43.37\% &40.72\% & 52.14\% \\
    
    \hline
    En$_2$ResNet20 (0.54M)
    & BR & \bf{72.58}\% & \bf{49.29}\% & \bf{44.72}\% &\bf{60.36}\%\\
    ResNet34 (0.56M)
    &BR&70.31\%&46.42\%&43.26\%&54.75\%\\
    \hline
    \end{tabular}
    \caption{Ensemble ResNet vs ResNets. We verify that EnResNet outperforms ResNet with the similar number of parameters with both float and binary weights.}
    \label{table 2}
\end{table}

\begin{table}[ht!]
    \centering
    \begin{tabular}{c c c c c c}
    \hline\hline
    Model & Quant &$N$ & $A_1$ & $A_2$ & $A_3$\\
    \hline
    \multirow{3}{*}{En$_1$ResNet20}&Float &78.31\% & 56.64\% & 49.00\%& 66.84\%\\
    &BC &68.84\% & 46.31\%&  42.45\%&58.52\%\\
    &BR &\bf{69.60}\% & \bf{47.17}\% & \bf{43.89}\% & \bf{58.79}\%\\
    \hline
    \multirow{3}{*}{En$_2$ResNet20}&Float &80.10\% &57.48\% & 49.55\%&66.73\% \\
    & BC & 71.48\% & 47.83\% & 43.03\%& 59.09\%\\
    & BR & \bf{72.58}\% & \bf{49.29}\% &\bf{ 44.72}\% &\bf{60.36}\%\\
    \iffalse
    \hline
    \multirow{3}{*}{En$_3$ResNet20}&Float& 80.99\%&57.53\%&49.61\% \\
    &BC&73.69\% & 49.52\% &45.01\% &59.75\%\\
    &BR&\bf{73.88}\%&\bf{50.06}\%&45.48\%&60.85\% \\
    \fi
    \hline
    \multirow{3}{*}{En$_5$ResNet20}&Float &80.64\% & 58.14\%& 50.32\%&66.96\%\\
    &BC &75.54\% &51.03\% & 46.01\%&60.92\%\\
    &BR & \bf{75.40}\%&\bf{51.60}\% & \bf{46.91}\% & \bf{61.52}\%\\
    \hline
    \end{tabular}
    \caption{Binary Connect vs Binary Relax. Models quantized by Binary-Relax have higher natural accuracy and adversarial accuracies than those quantized by Binary-Connect}
    \label{tab:binarize}
\end{table}

\section{Trade-off between robust accuracy and natural accuracy}
\subsection{Previous work and our methodology}
It is known that adversarial training will decrease the accuracy for classifying the clean input data. This phenomenon is verified both theoretically \cite{Dimitris} and experimentally \cite{Zhang, enresnet, Alexey} by researchers. \cite{Zhang} proposed a trade-off loss function \eqref{eq:trades} for robust training to balance the adversarial accuracy and natural accuracy. In
practice, it is formulated as following:
\begin{equation}
    \mathcal{L}=\mathcal{L}_{nat}+\beta\cdot \frac{1}{N}\sum_{n=1}^{N}l\big(f(x_n),f(\tilde{x}_n)\big)\label{eq:1}
\end{equation}
Motivated by \cite{Zhang}, we study the following trade-off loss function for our quantized models:
\begin{equation}
\mathcal{L}=\alpha\cdot\mathcal{L}_{nat}+\beta \mathcal\cdot{L}_{rob}\label{eq:2}
\end{equation}
Note that adversarial training is a special case $\alpha=0$, $\beta=1$ in ($\ref{eq:2}$).
Loss function (\ref{eq:1}), TRADES, improves the robustness of models by pushing the decision boundary away from original data points, clean images in this case. 
However, intuitively, if the model classifies a original data
point wrong, the second term of \eqref{eq:1} will still try to extend this
decision boundary, which can prevent the first term of the loss
function from leading the model to the correct classification.
In this section, we will experimentally compare \eqref{eq:1} and \eqref{eq:2}
and theoretically analyze the difference between them.
\begin{table}[ht!]
    \centering
    \begin{tabular}{c c c c c c}\\
    \hline
         Model& Loss &$N$ &$A_1$&$A_2$ &$A_3$\\
         \hline\hline
         En$_1$ResNet20& (\ref{eq:1}) ($\beta=1$)&\textbf{84.49}\%&45.96\%&34.81\%&51.94\%\\
         En$_1$ResNet20 & (\ref{eq:2}) ($\alpha=1,\beta=1$)&83.47\%&\textbf{54.46}\%&\textbf{43.86}\%&\textbf{64.04}\%\\
         \hline
         En$_1$ResNet20&(\ref{eq:1}) ($\beta=4$)& 80.05\% & 51.24\%&45.43\%&58.85\%\\
         En$_1$ResNet20& (\ref{eq:2}) ($\alpha=1,\beta=4$)&\textbf{80.91}\%& \textbf{55.92}\%& \textbf{47.17}\% &\textbf{66.53}\%\\
         \hline
         En$_1$ResNet20 & (\ref{eq:1}) ($\beta=8$)&75.82\%& 51.63\%&46.95\% &59.31\%\\
         En$_1$ResNet20 & (\ref{eq:2})  ($\alpha=1,\beta=8$)&\textbf{79.31}\%&\textbf{56.28}\%&\textbf{48.02}\%&\textbf{66.07}\%\\
                  \hline
    \end{tabular}
    \caption{Comparison of Loss Function. Trade-off loss function (\ref{eq:1}) outperforms TRADES (\ref{eq:2}) in most cases.}
    \label{tab:4}
\end{table}
\subsection{Experiment and result}
To compare the performances of two loss functions, we choose
%fix 
our neural network and dataset to be En$_1$ResNet20 and CIFAR-10 respectively. Based on \cite{Zhang}, who studied $\beta$ of \eqref{eq:1} in the range $[1,10]$, we vary the trade-off parameter $\beta$, the weight of adversarial loss, in the set $[1,4,8]$ to emphasize the robustness in different levels. \\
The experiment results are listed in Table \ref{tab:4}. We observe that, when the natural loss and the adversarial loss are equally treated, the model trained by \eqref{eq:1} has higher natural accuracy while \eqref{eq:2} makes its model more robust. As the trade-off parameter $\beta$ increases, natural accuracy of the model trained \eqref{eq:1} drops rapidly, while \eqref{eq:2} trades a relatively smaller amount of natural loss for robustness. As a result, when $\beta = 4$ and $\beta = 8$, 
the model trained by \eqref{eq:2} has both higher natural accuracy and higher adversarial accuracy.  
Hence, we say that \eqref{eq:2} has better trade-off efficiency than \eqref{eq:1}
\medskip
\subsection{Analysis of trade-off functions}
In this subsection, we present a theoretical analysis that why \eqref{eq:2} outperforms \eqref{eq:1}. Let us consider the binary classification case, where have our samples $(\mathbf{x},y) \in$ $\mathcal{X}\times \{-1,1\}$. Let $f: \mathcal{X}\rightarrow \mathbb{R}$ be a classifier and $\sigma(\cdot)$ be an activation function. Then our prediction for a sample is $\hat{y}=sign(f(\mathbf{x}))$ and the corresponding socre is $\sigma(f(\mathbf{x}))$. Above is the theoretical setting provided by \cite{Zhang}. Then, we have the errors $\mathcal{R}_{\phi}(f)$ and $\mathcal{R}_{\phi}^*(f)$ corresponding to \eqref{eq:1} and \eqref{eq:2} respectively:
$$\mathcal{R}_{\phi}(f)=\mathbb{E}[\phi(\sigma \circ f(\mathbf{x})\cdot y) ]+\beta \cdot \mathbb{E}[\phi(\sigma \circ f(\mathbf{x})\cdot \sigma \circ f(\mathbf{x}'))]$$
$$\mathcal{R}_{\phi}^*(f)=\alpha\cdot \mathbb{E}[\phi(\sigma \circ f(\mathbf{x})\cdot y) ]+\beta \cdot\mathbb{E}[\phi(\sigma \circ f(\mathbf{x}')\cdot y)]$$

\begin{table}[ht!]
    \centering
    \begin{tabular}{c@{\hspace{1.5\tabcolsep}} c@{\hspace{1.5\tabcolsep}} c@{\hspace{1.5\tabcolsep}} c@{\hspace{1.5\tabcolsep}} c@{\hspace{1.5\tabcolsep}} c}
    \hline
        Model & loss & $N$ & $A_1$ &$A_2$ & $A_3$\\
        \hline \hline
        En$_1$ResNet20 & $\alpha=0,\beta=1$ &69.60\%
        &\textbf{47.81\%} &\textbf{43.89\%} &58.79\% \\
        En$_1$ResNet20& $\alpha=1,\beta=4$&\textbf{73.40\%}&47.41\%&41.86\%&57.83\%\\
        En$_1$ResNet20 & $\alpha=1,\beta=8$& 71.35\% &47.42\% & 42.46\% &\textbf{59.01\%} \\
        \hline
        En$_2$ResNet20& $\alpha=0,\beta=1$ & 71.58\%&49.29\%& \textbf{44.62\%} & 60.36\%\\
        En$_2$ResNet20&$\alpha=1,\beta=4$& \textbf{75.92\%}& 48.97\%& 43.41\% &59.40\%\\
        En$_2$ResNet20& $\alpha=1,\beta=8$& 74.72\% &\textbf{49.66\%} &43.96\% & \textbf{60.65\%}\\
        \hline
        En$_5$ResNet20&$\alpha=0,\beta=1$&75.40\%&51.60\%&\textbf{46.91\%}&\textbf{61.52\%}\\
        En$_5$ResNet20&$\alpha=1,\beta=4$&\textbf{78.50\%}&50.85\%&45.02\%&60.96\%\\
        En$_5$ResNet20& $\alpha=1,\beta=8$&77.35\%&\textbf{51.62\%}&45.63\%&61.11\%\\
        \hline
    \end{tabular}
    \caption{Trade-off loss function for binarized models with different parameters. When $\alpha=1$ and $\beta=8$, models can achieve about the same accuracies under FGSM and C\&W attacks as solely adversarially trained models, while the natural accuracy is improved.}
    \label{tab:5}
\end{table}

Now, we consider several common loss functions: the hinge loss ($\phi(\theta)=\max\{1-\theta,0\}$), the sigmoid loss ($\phi(\theta)=1-\tanh{\theta}$), and the logistic loss ($\phi(\theta)=\log_2(1+e^{-\theta})$). Note that we want a loss function to be monotonically deceasing in the interval $[-1,1]$ as $-1$ indicates that the prediction is completely wrong, and $1$ indicates the prediction is completely correct.  Since our classes is 1 and -1, we will choose hyperbolic tangent as our activation function.\\

\begin{figure}
    \centering
    \subfloat[Ternary AT]{\includegraphics[scale=0.4]{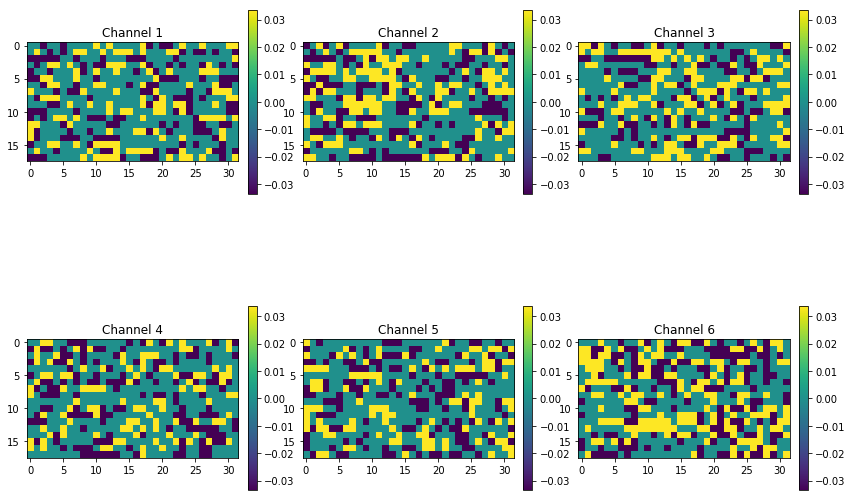}}
    % \qquad
    
    \subfloat[Ternary Trade-off]{\includegraphics[scale=0.4]{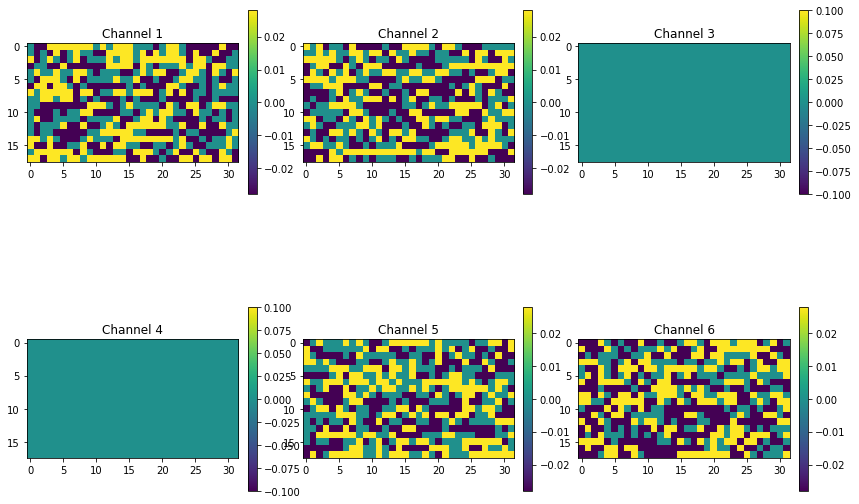}}
    \caption{Visualization of 4 ternary channels (reshaped for visulization) of a layer in En$_1$ResNet56 under natural training. There are less 0 weights in non-sparse channels. Nonzero weights of ternary channels under natural training  are more concentrated. As a result, there are more sparse channels under natural training.}
    \label{fig:tnn}
\end{figure}
\begin{table*}[ht!]
    \centering
    \begin{tabular}{c c c c c c c c}\hline
        Model & Quant &Loss & Weight Sparsity & Channel Sparsity &$N$ & $A_1$ &$A_2$\\
        \hline \hline
        ResNet20 & tnn & Natural &53.00\%
        &11.16\% & 90.54\%&12.71\%&0.00\% \\
        En$_1$ResNet20& tnn & Natural &52.19\%&9.57\%&90.61\%&26.21\%&0.71\%\\
        ResNet20 & tnn & AT & 50.71\% &2.55\% & 68.30\% &44.80\% &42.53\%\\
        En$_1$ResNet20 & tnn & AT&50.31\%&\cellcolor{yellow}4.14\%&71.30\%&48.17\%&43.27\%\\
        En$_1$ResNet20 & tnn &  (\ref{eq:2})&55.66\%&\cellcolor{green}7.02\%&73.05\%&48.10\%&42.65\%\\
        \hline
        ResNet20 & 4-bit & Natural &42.79\%
        &9.53\% &91.75\% &12.38\%&0.00\% \\
        En$_1$ResNet20& 4-bit & Natural &44.73\%&10.52\%&91.42\%&27.99\%&0.62\%\\
        ResNet20 & 4-bit & AT & 43.93\% &2.55\% & 71.49\% &47.63\% &44.08\%\\
        En$_1$ResNet20 & 4-bit & AT&48.35\%&\cellcolor{yellow}4.94\%&73.05\%&51.43\%&45.10\%\\
        En$_1$ResNet20 &4-bit & \eqref{eq:2}& 55.57\%&\cellcolor{green}7.42\%&76.61\%&51.92\%&44.39\%\\
        \hline
         ResNet56&tnn&Natural&60.96\%&31.86\%&91.91\%&15.58\%&0.00\%\\
         En$_1$ResNet56&tnn&Natural&60.66\%&28.97\%&91.46\%&38.22\%&0.36\%\\
         ResNet56&tnn&AT&54.21\%&15.37\%&74.56\%&51.73\%&46.62\%\\
         En$_1$ResNet56&tnn&AT&54.70\%&\cellcolor{yellow}16.74\%&76.87\%&53.16\%&47.89\%\\
         En$_1$ResNet56&tnn&(\ref{eq:2})&58.89\%&\cellcolor{green}21.36\%&77.24\%&52.96\%&46.01\%\\
        \hline
         ResNet56&4-bit&Natural&67.94\%&39.16\%&93.09\%&16.02\%&0.00\%\\
         En$_1$ResNet56&4-bit&Natural&71.07\%&41.10\%&92.39\%&39.79\%&0.33\%\\
         ResNet56&4-bit&AT&55.29\%&17.10\%&77.67\%&52.43\%&48.22\%\\
         En$_1$ResNet56&4-bit&AT&55.09\%&\cellcolor{yellow}18.11\%&78.25\%&55.48\%&49.03\%\\
         En$_1$ResNet56&4-bit&(\ref{eq:2})&67.31\%&\cellcolor{green}33.18\%&79.44\%&55.41\%&47.80\%\\
         \hline
    \end{tabular}
    \caption{Sparsity and structure of sparsity of quantized models. We use $\alpha=1$ and $\beta=8$ in trade-off loss (\ref{eq:2}). While models under both natural training and adversarial have large proportion of sparse weights, sparsity of adversarially trained models are much less structured. Trade-off loss function (\ref{eq:trades}) can improve the structure.}
    \label{tab:7}
\end{table*}

\begin{proposition}
Let $\phi$ be any loss function that is monotonically decreasing on $[-1,1]$ (all loss functions mentioned above satisfy this), and $\sigma(\theta)=\tanh{\theta}$. Define $B=\{\mathbf{x}|f(\mathbf{x})y\geq 0, f(\mathbf{x'})y\geq 0\}$. Then:
$$\mathcal{R}_{\phi}(f)\geq \mathcal{R}_{\phi}^*(f) \text{ on } B \text{  and  }\mathcal{R}_{\phi}(f)\leq \mathcal{R}_{\phi}^*(f) \text{ on }B^C$$
Proof: 
%Let $E$, $D$, and $F$ be the sets defined in proposition 1. 
We first define a set $E$:
$$E=\{\mathbf{x}|f(\mathbf{x})y<0,f(\mathbf{x'})y>0\}$$
By the definition of adversarial examples in \eqref{eq:ad}, $$\mathbb{E}[{\mathbf{1}\{E\}}]=0$$
where $\mathbf{1}\{E\}$ is the indicator function of the set. That is, the set that the classifier predicts original data point wrong but the perturbed data point correctly should have measure $0$. Now, we define the following sets:
$$D=\{\mathbf{x}|f(\mathbf{x})y\geq 0, f(\mathbf{x'})y< 0\} $$
$$F=\{\mathbf{x}|f(\mathbf{x})y< 0, f(\mathbf{x'})y<0\}$$
We note that the activation function $\sigma(x)$ preserves the sign of $x$, and $|\sigma(x)|\leq 1$.\\
On the set $B$, $f(\mathbf{x})$, $f(\mathbf{x'})$, and $y$ have the same sign, so are $\sigma(f(\mathbf{x}))$, $\sigma(f(\mathbf{x'}))$ and $y$. Therefore $$\phi(\sigma(f(\mathbf{x'}))\cdot \sigma(f(\mathbf{x}))\geq \phi(\sigma(f(\mathbf{x'})\cdot y)$$ as $0\leq \sigma( f(\mathbf{x'}))\cdot \sigma(f(\mathbf{x}))\leq \sigma(f(\mathbf{x'}))\cdot y$. This shows 
$$\mathcal{R}_{\phi}(f)\geq \mathcal{R}_{\phi}^*(f) \text{ on }B$$
We note that $B^C=E \cup D \cup F$. Since set $E$ has measure zero, we only consider $D$ and $F$. \\ 
On $D$, as $f$ classifies $\mathbf{x}$ correct and $\mathbf{x'}$ wrong, we have 
$$ \sigma(f(\mathbf{x'}))\cdot y \leq \sigma(f(\mathbf{x'}))\cdot \sigma(f(\mathbf{x}))\leq0$$
$$\Rightarrow \phi(\sigma(f(\mathbf{x'}))\cdot \sigma(f(\mathbf{x}))\leq \phi(\sigma(f(\mathbf{x'})\cdot y)$$
On $F$, as $f$ classifies both $\mathbf{x}$ and $\mathbf{x'}$ wrong, we have 
$$ \sigma(f(\mathbf{x'}))\cdot y \leq 0 \leq \sigma(f(\mathbf{x'}))\cdot \sigma(f(\mathbf{x}))$$
$$\Rightarrow \phi(\sigma(f(\mathbf{x'}))\cdot \sigma(f(\mathbf{x}))\leq \phi(\sigma(f(\mathbf{x'})\cdot y)$$
In summary, we have 
$$\mathcal{R}_{\phi}(f)\leq \mathcal{R}_{\phi}^*(f) \text{ on  } B^C$$
\hfill $\square$\\
\end{proposition}

We partition our space into several sets based on a given classifier $f$, and we examine the actions of loss functions on those sets. We see that \eqref{eq:1} penalize set $B$ heavier than \eqref{eq:2}, but the classifier classifies both the natural data and the perturbed data correct on $B$. On the other hand, \eqref{eq:1} does not penalize sets $E$ and $F$, where the classifier makes mistakes, enough, especially on set $F$. Therefore, \eqref{eq:2} as a loss function is more on target. Based on both experimental results and theoretical analysis, we believe \eqref{eq:2} is a better choice to balance natural accuracy and robust accuracy.

As our experiments on the balance of accuracies with different parameters in our loss function (\ref{eq:2}) in table \ref{tab:5}. We find that it is possible to increase the natural accuracy while maintaining the robustness under relatively weak attacks (FGSM \& CW), as the case of $\alpha=1$ and $\beta=8$ in table \ref{tab:5}. However, the resistance under relatively strong attack (IFGSM) will inevitably decrease when we trade-off.
\iffalse
\begin{figure}
    \centering
    \subfloat[4-bit AT]{\includegraphics[scale=0.2]{4bit2.png}}
    \subfloat[4-bit AT]{\includegraphics[scale=0.2]{4bit1.png}}
    \qquad
    \subfloat[4-bit Trade-off]{\includegraphics[scale=0.2]{fbit_trades2.png}}
    \subfloat[4-bit Trade-off]{\includegraphics[scale=0.2]{fbit_trades1.png}}
    \caption{Comparison of Adversarial training (AT) with trade-off loss: the left-hand side are original plots of channels, and we isolate the zero weights and nonzero weights of the left-hand side channels for better visualization in right-hand side. The dark parts are zero weights and the yellow parts are nonzero weights. We observe that the trade-off loss concentrates nonzero weights into non-sparse channels comparing to the adversarial loss }
\end{figure}
\fi
\section{Further balance of efficiency and robustness: Structured sparse quantized neural network via Ternary/4-bit quantization and trade-off loss function}
\subsection{Sparse neural network delivered by high-precision quantization}
When we quantize DNNs with precision higher than binary, such as ternery and 4-bit quantization, zero is in quantization levels. In fact, we find that a large proportion of weights will be quantized to zero. This suggests that a ternary or 4-bit quantized model can be further simplified via channel pruning. However, such simplification requires structure sparsity of DNN architecture. In our study, we use the algorithm 1 as before with the projection replaced by ternary and  4-bit respectively. As shown in Table \ref{tab:7}, we find that sparsity of quantized DNNs under regular training are significantly more structured than those under adversarial training. For both ternary and 4-bit quantization, quantized models with adversarial training have very unstructured sparsity, while models with natural training have much more structured sparsity. For example, 50.71\% (0.135M out of 0.268M) of weights in convolutional layers are zero in a ternary quantized ResNet20, but there are only 2.55\% (16 out of 627) channels are completely zero. If the sparsity is unstructured, it is less useful for model simplification as channel pruning cannot be applied. A fix to this problem is our trade-off loss function, as factor natural loss into adversarial training should improve the structure of sparsity. Our experiment (Table \ref{tab:7}) shows that a small factor of natural loss, $\alpha=1$ and $\beta$ in (\ref{eq:2}), can push the sparsity to be more structured. Meanwhile, the deepness of models also has an impact on the structure of sparsity. The deeper the more structured the sparsity is. We see in Table \ref{tab:7} that, under the same settings, the structure of sparsity increases as the model becomes deeper. 
Figure \ref{fig:tnn} shows the difference between a unstructured sparsity of a ternary ResNet20 with adversarial training and a much more structured sparsity of ResNet56 with natural training. The trade-off function not only improves the natural accuracy of models with merely minor harm to robustness but also structures the sparsity of high-precision quantization, so further simplification of models can be done through channel pruning.
\begin{table*}[]
    \centering
    \begin{tabular}{c c c c c c c c c c}
    \hline
        Model & Size & Dataset & Loss & Quant & Ch Sparsity & N &$A_1$&$A_2$&$A_3$ \\
        \hline
        \hline
         En$_2$ResNet20& 0.54M& MNIST& (\ref{eq:2}) &BR& N/A &99.22\%&98.90\%&98.90\%&99.12\% \\
         
         ResNet44& 0.66M & MNIST& TRADES&Float&N/A &99.31\%&98.98\%&98.91\%&99.14\%\\
         En$_2$ResNet20& 0.54M& MNIST& (\ref{eq:2}) &Float& N/A &99.21\%&99.02\%&98.91\%&99.14\% \\
         
         \hline
         En$_2$ResNet20 & 0.54M &FMNIST& (\ref{eq:2})& BR &N/A& 91.69\%&87.85\%&87.22\%&89.74\%\\
         
         ResNet44 & 0.66M & FMNIST& TRADES &Float &N/A& 91.37\%&88.13\%&87.98\%&90.12\% \\
         En$_2$ResNet20 & 0.54M & FMNIST& (\ref{eq:2}) &Float &N/A& 92.74\%&89.35\%&88.68\%&91.72\%\\
         \hline
         En$_1$ResNet56&0.85M&Cifar10& (\ref{eq:2})&4-bit&33.18\%&79.44\%&55.71\%&47.81\%&65.50\%\\
         ResNet56&0.85M&Cifar10&TRADES&Float&N/A&78.92\%&55.27\%&50.40\%&59.48\%\\
         En$_1$ResNet56&0.85M&Cifar10& (\ref{eq:2})&Float&N/A&81.63\%&56.80\%&50.17\%&66.56\%\\
         \hline
         En$_1$ResNet110 & 1.7M & Cifar100 &(\ref{eq:2}) &4-bit&23.93\% &53.08\%&30.76\%&25.73\%&42.54\%\\
         ResNet110&1.7M & Cifar100&TRADES &Float&N/A&51.65\%&28.23\%&25.77\%&40.79\%\\
         En$_1$ResNet110 & 1.7M & Cifar100 &(\ref{eq:2}) &Float&N/A &56.63\%&32.24\%&26.72\%&43.99\%\\
         \hline
         En$_2$ResNet56& 1.7M &SVHN&(\ref{eq:2})&4-bit&49.03\% &91.21\%&70.91\%&57.99\%&72.44\%\\
         ResNet110& 1.7M& SVHN&TRADES& Float&N/A&88.33\%&64.80\%&56.81\%&69.79\%\\
         En$_2$ResNet56& 1.7M &SVHN&(\ref{eq:2})&Float&N/A &93.33\%&78.08\%&59.11\%&75.79\%\\
         \hline
    \end{tabular}
    \caption{Generalization of quantized models to more datasets}
    \label{tab:6}
\end{table*}

\section{Benckmarking Adversarial Robustness of Quantized Model}
\iffalse
\begin{figure}[ht!]
    \centering
    \subfloat[SVHN]{\includegraphics[scale=0.7]{SVHN_batch.png}}
    \qquad
    \subfloat[Fashion-MNIST]{\includegraphics[scale=0.6]{FMNIST_batch.png}}
    \caption{a mini-batch of SVHN dataset and a mini-batch of Fashion-MNIST dataset}
    \label{fig:batch}
\end{figure}
\fi
Based on previous discussions, integrating the relaxation algorithm, ensemble ResNet, and our trade-off loss function, we can produce very efficient DNN models with high robustness. To our best knowledge, there is no previous work that systematically study the robustness of quantized models. As a result, we do not have any direct baseline to measure our results. Therefore, we benchmark our results by comparing to models with similar size and current state-of-the-art defense methods. We verify that the performance of the quantized model with our approach on popular datasets, including Cifar 10, Cifar 100 \cite{cifar}, MNIST, Fashion MNIST (FMNIST) \cite{FMNIST}, and SVHN \cite{SVHN}. In our experiments, we learn SVHN dataset without utilizing its extra training data. Our results are displayed in Table \ref{tab:6}. In this table, size refers to the number of parameters. We find that quantization has very little impact on learning small datasets MNIST and FMNIST. En$_2$ResNet20 and ResNet40 have about the same performance while the previous is binarized. In fact, a binary En$_2$ResNet20 has about the same performance as its float equivalent, which means we get efficiency for 'free' on these small datasets.  We benchmark the robustness of qunaitzed models on large datasets, SVHN, Cifar10, and Ciar100, using models with 4-bit quantization. As in Table \ref{tab:6}, 4-bit EnResNets with loss function \eqref{eq:2} have better performance than float TRADES-trained ResNets with the same sizes. However, quantized models on these larger datasets are outperformed by their float equivalents. In another word, efficiency of models are not 'free' when learning large datasets, we have to trade-off between performance and efficiency. Although 4-bit quantization requires higher precision and more memories, the highly structured sparsity can compensate the efficiency of models. The 4-bit quantized models of En$_1$ResNet56 for Cifar10, En$_1$ResNet110 for Cifar 100, and En$_2$ResNet56 for SVNH in Table \ref{tab:6} have sizes of 0.58M, 1.43M, and 0.80M respectively if the sparse channels are pruned. Our codes as well as our trained quantized models listed in Table \ref{tab:6} are available at
\href{https://github.com/lzj994/Binary-Quantization}{https://github.com/lzj994/Binary-Quantization}.

\section{Conclusion}
In this paper, we study the robustness of quantized models. The experimental results suggest that it is totally possible to acheive both efficiency and robustness as quantized models can also do a good job at resisting adversarial attack. Moreover, we discover that adversarial training will make the sparsity from high-precision quantization unstructured, and a trade-off function can improve the sparsity structure. With our integrated approach to balance efficiency and robustness, we find that keeping a model both efficient and robust is promising and worth paying attention to. We hope our study can serve as a benchmark for future studies on this interesting topic. 

%
% the environments 'definition', 'lemma', 'proposition', 'corollary',
% 'remark', and 'example' are defined in the LLNCS documentclass as well.
%

\bibliographystyle{plain}
\bibliography{egbib}

\end{document}